%% file: 0.main.tex
\documentclass[11pt]{article}
\usepackage{coling2020}
\usepackage{times}
\usepackage{url}
\usepackage{latexsym}

\usepackage{graphicx}
\usepackage{amsmath}
\usepackage{multirow}
\usepackage{booktabs}
\usepackage{color, xcolor}
\definecolor{Gray}{gray}{0.9}
\definecolor{caseblue}{rgb}{0.27451,0.5098,0.70588}
\definecolor{casered}{rgb}{0.27451,0.5098,0.70588}
\usepackage{amssymb}
\usepackage{enumitem}
\usepackage{microtype}
\usepackage{relsize}
\usepackage{stmaryrd}
\usepackage{amsfonts}
\usepackage{soul}
\usepackage{arydshln}
\usepackage{tabularx}
\newcommand{\dashrule}[1][black]{%
  \color{#1}\rule[\dimexpr.5ex-.2pt]{4pt}{.4pt}\xleaders\hbox{\rule{4pt}{0pt}\rule[\dimexpr.5ex-.2pt]{4pt}{.4pt}}\hfill\kern0pt%
}

\colingfinalcopy 

\title{EmpDG: Multi-resolution Interactive Empathetic Dialogue Generation}

\author{Qintong Li$^{1}$, Hongshen Chen$^{2}$, Zhaochun Ren$^{1,}$\thanks{{} {} Corresponding author.}, Pengjie Ren$^{1}$, Zhaopeng Tu$^{3}$,Zhumin Chen$^{1}$\\
	$^{1}$School of Computer Science and Technology, Shandong University, Qingdao, China \\
	$^{2}$Data Science Lab, JD.COM, Beijing, China\\
  $^{3}$Tencent AI Lab, Shenzhen, China\\
	{\tt qintongli@mail.sdu.edu.cn, ac@chenhongshen.com,} \\ 
	{\tt \{zhaochun.ren, chenzhumin\}@sdu.edu.cn,} \\ 
	{\tt jay.ren@outlook.com, zptu@tencent.com}
}

\date{}

\begin{document}
\maketitle
\begin{abstract}
  A humanized dialogue system is expected to generate empathetic replies, which should be sensitive to the users' expressed emotion. 
  The task of empathetic dialogue generation is proposed to address this problem.
  The essential challenges lie in accurately capturing the nuances of human emotion and considering the potential of user feedback, which are overlooked by the majority of existing work. In response to this problem, we propose a multi-resolution adversarial model -- EmpDG, to generate more empathetic responses. EmpDG exploits both the coarse-grained dialogue-level and fine-grained token-level emotions, the latter of which helps to better capture the nuances of user emotion. 
  In addition, we introduce an interactive adversarial learning framework which exploits the user feedback, to identify whether the generated responses evoke emotion perceptivity in dialogues.
  Experimental results show that the proposed approach significantly outperforms the state-of-the-art baselines in both content quality and emotion perceptivity.
\end{abstract}

\input{1.introduction}

\input{2.related_work}

\input{3.model}
\input{4.experiment}
\input{5.conclusion}

\bibliographystyle{coling}
\bibliography{coling2020}

\end{document}

%% file: 1.introduction.tex
\section{Introduction}

\blfootnote{
	This work is licensed under a Creative Commons 
	Attribution 4.0 International License.
	 License details:
	\url{http://creativecommons.org/licenses/by/4.0/}
}

Studies on social psychology suggest that ``empathy'' is a crucial step towards a more humanized  human-machine conversation, which improves the emotional perceptivity in emotion-bonding social activities~\cite{zech2005talking}.
To design an intelligent automatic dialogue system, it is important to make the chatbot become empathetic within dialogue interactions~\cite{prendinger2005empathic}. 
Therefore, in this paper, we focus on the task of \textit{empathetic dialogue generation}~\cite{rashkin2019towards}, which automatically tracks and understands the user emotion information in a multi-turn dialogue scenario.

\begin{figure}[!t]
    \centering
    \includegraphics[width=0.88\textwidth]{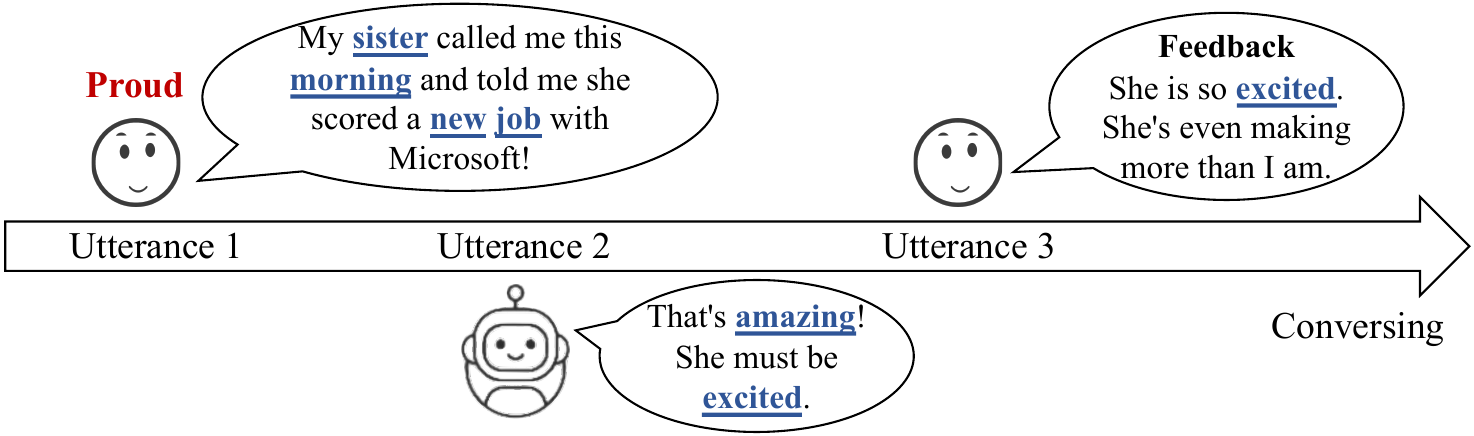}
    \caption{An empathetic dialogue example from dataset \textsc{EmpatheticDialogues}. The emotional-related words are highlighted in \underline{\textbf{\textcolor{caseblue}{blue}}}. ``Proud'' is the coarse-grained emotional label.  These emotional words are labelled by an external emotion lexicon~\cite{mohammad2013crowdsourcing}.}  
    \label{fig:example}
\end{figure}

Despite the achieved successes~\cite{rashkin2019towards,LinMSXF19}, obstacles to establishing an empathetic conversational system are still far beyond current progress:
(1) It is still difficult to accurately capture the nuances of human emotion in dialogue generation~\cite{ghosal2019dialoguegcn}.
(2) Merely relying on the dialogue history but overlooking the potential of user feedback for the generated responses further aggravates the aforementioned deficiencies, which causes undesirable responses~\cite{zhang2018exploring}.
In Figure~\ref{fig:example}, we give an example of the benchmark dataset \textsc{EmpatheticDialogues}~\cite{rashkin2019towards}.
Notably, the emotional words in the utterance 1~(i.e., ``new'', ``job'') and utterance 2~(i.e., ``amazing'', ``excited'') have fine-grained emotional connections.
Without considering fine-grained emotional words, the responses generated by existing methods are trivial and uninformed, even though they expressed the appropriate emotions~(e.g., ``That's great.' ). 
Moreover, we see the user’s feedback~(utterance 3) also has a close emotional connection to the target response~(utterance 2).
Therefore, explicitly modelling the fine-grained emotional factor and considering user feedback are necessary.

In this paper, we propose a multi-resolution adversarial empathetic dialogue generation model, named EmpDG, to address above challenges through generating more appropriate and empathetic responses. 
To capture the nuances of user emotions sufficiently, EmpDG generates responses by jointly taking coarse-grained dialogue-level emotions and fine-grained token-level emotions into account.
The multi-resolution emotions are the prerequisites of the response generator to perceive the user's nuanced emotion states.
Furthermore, we propose an interactive adversarial learning framework to augment the user's feedback thoughtfully, where two interactive discriminators are designed to identify whether the generated responses evoke the emotion perceptivity regarding both the dialogue history and the user's emotions.
Conducted on a benchmark dataset \textsc{EmpatheticDialogues}, extensive experimental results have verified the effectiveness of the EmpDG in terms of both content quality and empathy quality. We also find that the EmpDG outperforms state-of-the-art baselines significantly.

\smallskip\noindent
\noindent In general, our main contributions are summarized as follows:
\begin{itemize}
\item We propose a multi-resolution adversarial neural network, which considers multi-granularity emotion factors in the dialogue context.
\item To induce the response generation from users' feedback, we propose an interactive adversarial learning network with two interactive discriminators. 
\item  Experiments show that EmpDG significantly outperforms state-of-the-art baselines in terms of both content quality and empathy quality in the empathetic dialogue generation.
\end{itemize}

%% file: 2.related_work.tex
\section{Related Work}
Our research is in line with empathetic conversation generation through human-computer interactions, which avoids the additional step of assigning emotion to responses during conversations~\cite{skowron2013affect}.
Several work~\cite{lubis2018eliciting,Rashkin18,zhong2019affect,wei2019emotion,Shin19,ChatterjeeGCSGA19,rashkin2019towards,Santhanam19,LinMSXF19,zhong2020endowing,lin2020caire} have attempted to make dialogue models more empathetic.
Rashkin et al.~\shortcite{rashkin2019towards} combine existing models in different ways to produce empathetic responses.
Lin et al.~\shortcite{LinMSXF19} softly combine the possible emotional responses from several separate experts to generate the final empathetic response.
Lin et al.~\shortcite{lin2020caire} fine-tune a large-scale pre-trained language model with multiple objectives.
However, all existing approaches only consider monogranular emotions in the dialogue context.
In comparison, EmpDG jointly considers highly correlated coarse-grained emotional labels and fine-grained emotional terms.
Moreover, we explicitly consider the effect of user feedback via a novel interactive adversarial mechanism, so that EmpDG is capable to evoke more emotion perceptivity in dialogues.

Besides the advancements in empathetic dialogue models, the emergence of new emotion-labelled dialogue corpora, e.g., \textsc{DailyDialog}~\cite{li2017dailydialog} and \textsc{EmotionLines}~\cite{HsuCKHK18}, have also contributed to this research field.
However, only 5\% of the utterances in \textsc{DailyDialog} and 16.68\% of the utterances in \textsc{EmotionLines} have diverse emotional labels and others are either “none” or “happy” labels. 
Because of the extremely imbalanced data distribution, they are not suitable to be engaged as the benchmarks of empathetic dialogue generation.
Rashkin et al.~\shortcite{rashkin2019towards} consider a richer and evenly distributed set of emotions and release a dataset \textsc{EmpatheticDialogues}, where a listener responds to a speaker who is under an emotional situation in an empathetic way.
Furthermore, several emotion lexicons~\cite{mohammad2013crowdsourcing,SedocBNBU20} have also been shown to be effective in tracking emotions in texts.
In our work, we focus on the task of empathetic dialogue generation on the \textsc{EmpatheticDialogues} dataset and emotion lexicon~\cite{mohammad2013crowdsourcing}.

The second line of our related work is the emotional dialogue generation, which has received an increasing amount of attention to address emotion factors~\cite{zhou2018emotional,huang2018automatic,colombo2019affect}.
Prior studies on emotion-related conversational systems mainly focused on rule-based systems, which heavily rely on hand-crafted features~\cite{prendinger2005empathic}. 
Recently, many neural emotional dialogue generation approaches~\cite{ghosh2017affect,zhou2018mojitalk,zhou2018emotional,huang2018automatic,colombo2019affect} have been explored to control the emotional expression in the target response.
However, as Li et al.~\shortcite{li2018syntactically} reveal that conventional emotional conversation systems aim to produce more emotion-rich responses according to a given specific user-input emotion, which inevitably leads to an emotional inconsistency problem. 

Our research also aligns with recent advances in open-domain dialogue generation models~\cite{vinyals2015neural,li2016deep,ZhangCWZLZL18,HancockBMW19,LiLBLL20,SongWZLL20}.
These dialogue models usually adopt the sequence-to-sequence (Seq2Seq)~\cite{sutskever2014sequence} fashion.
Adversarial learning achieves considerable success in generating higher-quality responses~\cite{goodfellow2014generative,li2017adversarial} but often leads to gradient vanishing as the discriminator saturates~\cite{gulrajani2017improved}. 
To tackle this problem, 
Gao et al.~\shortcite{gao2019product} utilize Wasserstein GAN~\cite{arjovsky2017wasserstein} to enhance response consistency with external facts.
Romanov et al.~\shortcite{RomanovRRD19} propose an adversarial decomposition method for fine-grained text representation.
Unlike the previous work, we investigate an adversarial approach to improve the empathy quality of neural dialogue models.

%% file: 3.model.tex
\section{Problem Formulation}

Before detailing our method, we introduce our key notations and concepts.
A multi-turn dialogue context consists of $M$ utterances between two interlocutors.
We assume both \emph{semantic context} and \emph{emotional context} exist in such dialogue.
The semantic context $\mathcal{U}$ refers to the sequence of utterances, i.e., $\mathcal{U}=[U_1, ..., U_M]$.
Following Lin~\shortcite{LinMSXF19}, we flat $\mathcal{U}$ into a long token sequence and insert a \texttt{CLS} token at the start of the token sentence, i.e., $\mathcal{U} = [\texttt{CLS}, x_1, \dots, x_m]$.
The emotional context $\mathcal{E}$ considers emotions with different granularities, i.e., $\mathcal{E}=[\texttt{LAB}, w_1,..., w_e]$, where $w_i$ is the emotional word in the semantic context $\mathcal{U}$, $\texttt{LAB}$ is a special emotion token which is used to derive the emotional state of the dialogue context.
We extract emotional words using an external emotional vocabulary $V_E$~\cite{mohammad2013crowdsourcing}.
In the following part, we use $x_0$ to denote \texttt{CLS} in $\mathcal{U}$ and $w_0$ to denote \texttt{LAB} in $\mathcal{E}$.

Given $\mathcal{U}$ and $\mathcal{E}$, our model aims to generate a $n$-length response $\mathcal{Y}=\{y_1, ..., y_n\}$ through maximizing the probability $P(\mathcal{Y}|\mathcal{U},\mathcal{E})=\prod_{n=1}^N P(y_n|y_{< n},\mathcal{U}, \mathcal{E})$.

\section{EmpDG}
\label{sec3}

\begin{figure*}[t]
 \centering
  \includegraphics[scale=0.53]{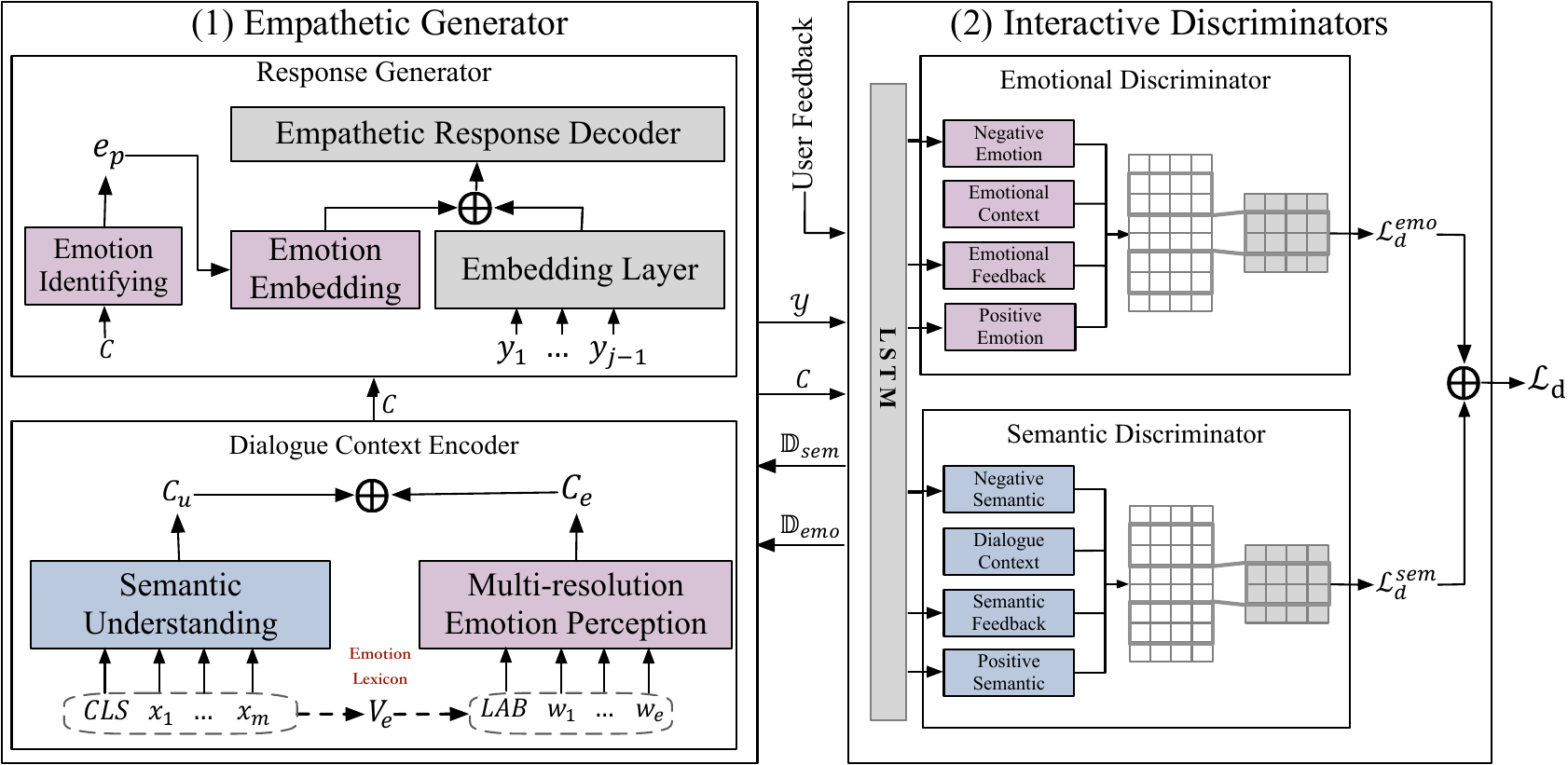}
  \caption{Overview of EmpDG. We divide EmpDG into two parts: (1) \textit{Empathetic generator} generates empathetic responses based on the semantic understanding and multi-resolution emotion perception. (2) \textit{Interactive discriminators} distinguishes whether the generated responses are emotion-appropriate and context-consistent.}
 \label{model}
\end{figure*}

In this section, we detail our proposed multi-resolutional adversarial model, abbreviated as EmpDG.
The overview of EmpDG is illustrated in Figure \ref{model}. There are two main components in EmpDG: the \textit{empathetic generator} and the \textit{interactive discriminators}.

To summarize, the \textit{empathetic generator} is established based on an encoder-decoder architecture, which are all implemented with Transformer~\cite{vaswani2017attention}.
During encoding procedure, semantic context and the multi-resolution emotional context are encoded; whereas the decoder fuses the semantic context and emotional context to generate responses. 

To enhance the empathy of the generator, we design two CNN-based~\cite{KalchbrennerGB14} discriminators (i.e., the \textit{semantic discriminator} and the \textit{emotional discriminator}).
In the training procedure, the two discriminators additionally interact with the user feedback (the next utterance and corresponding emotional words of the next utterance).
By minimizing the Wasserstein-1 distance to optimize the discriminators, we use the sum of classification results as a training signal to encourage response generator to evoke more emotion perceptivity.

\subsection{Empathetic Generator}
\label{ssec:smantic}

We start by proposing an empathetic generator to generate a response $\mathcal{Y}$. 
To better model the multi-granularity emotions, we divide our encoding-decoding process into $3$ phases individually: semantic understanding, multi-resolution emotion perception, and empathetic response generation.

\paragraph{Semantic Understanding.}
We first use a word embedding layer and a positional embedding layer~\cite{vaswani2017attention} to convert each token of semantic context $\mathcal{U}$ into vector representations $\textbf{e}^W_{x_i} \in \mathbb{R}^d$ and $\textbf{e}^P_{x_i} \in \mathbb{R}^d$, where $d$ is the dimensionality of embeddings.
In the multi-turn dialogue settings, distinguishing utterances from speaker and listener is helpful. Therefore, we incorporate the dialogue state embedding $\textbf{e}^D_{x_i}$ for each token. 
Our semantic context embedding $[\mathbf{x}_0,\ldots,\mathbf{x}_m]$ is the composition of three types of embeddings, where $\mathbf{x}_i$ is computed as follows:
\begin{equation}
\setlength{\abovedisplayskip}{3pt}
\setlength{\belowdisplayskip}{3pt}
    \mathbf{x}_i = \textbf{e}^W_{x_i} + \textbf{e}^P_{x_i} + \textbf{e}^D_{x_i}.
\end{equation}
Then we use the local transformer layer~\cite{vaswani2017attention} to encode semantic context information for each tokens $\mathbf{x}_i$. 
\begin{eqnarray}
\setlength{\abovedisplayskip}{3pt}
\setlength{\belowdisplayskip}{3pt}
    \mathbf{h}_i &=& \textrm{LayerNorm}(\mathbf{x}_i^{l-1} + \textrm{MHAtt}(\mathbf{x}_i^{l-1})), \\
    \mathbf{\tilde{x}}_i^l &=& \textrm{LayerNorm}(\mathbf{h}_i + \textrm{FFN}(\mathbf{h}_i)),
\end{eqnarray} 
where \textrm{LayerNorm} is the Layer Normalization trick proposed in~\cite{BaKH16}; \textrm{FFN} is a two-layer feed-forward network with ReLU as hidden activation function.
The transformer layers are stacked $L$ times. The obtained final context representations are denoted as $\mathbf{C_u} = [\mathbf{\tilde{x}}_0,\ldots, \mathbf{\tilde{x}}_m]$.

\paragraph{Multi-resolution Emotion Perception.}
We use another transformer encoder with a different set of parameters to encode the multi-resolution emotional context $\mathcal{E}$.
\begin{eqnarray}
\setlength{\abovedisplayskip}{3pt}
\setlength{\belowdisplayskip}{3pt}
    \mathbf{w}_i &=& \textbf{e}^W_{w_i} + \textbf{e}^P_{w_i} + \textbf{e}^E_{w_i}, \\
    \mathbf{h}_i &=& \textrm{LayerNorm}(\mathbf{w}_i^{l-1} + \textrm{MHAtt}(\mathbf{w}_i^{l-1})), \\
    \mathbf{\tilde{w}}_i^l &=& \textrm{LayerNorm}(\mathbf{h}_i + \textrm{FFN}(\mathbf{h}_i)),
\end{eqnarray} 
where $\textbf{e}^E_{w_i}$ is the emotion state embedding to distinguish the emotional context from semantic context. Then the multi-resolution emotional context is represented as $\mathbf{C_e}=[\mathbf{\tilde{w}}_0,\ldots,\mathbf{\tilde{w}}_e]$.

To perceive the emotional information in dialogue context, a linear layer with $\operatorname{softmax}$ operation projects the concatenation of $\mathbf{\tilde{w}}_0$ and $\mathbf{\tilde{x}}_0$ into an emotion category distribution $P_e$ over the coarsed-grained emotional label $e$ to identify the emotion signal user expressed:
\begin{eqnarray}
\setlength{\abovedisplayskip}{3pt}
\setlength{\belowdisplayskip}{3pt}
    \mathbf{e}_p = &\mathbf{W}_{e}[\mathbf{\tilde{w}}_0;\mathbf{\tilde{x}}_0],\\
    P_e(e|\mathcal{E}) = &\textrm{\textrm{softmax}}(\mathbf{e}_p),
\end{eqnarray}
where $\mathbf{W}_{e} \in \mathbb{R}^{2d \times q}$, $q$ is the number of emotion categories, and $[;]$ is the concatenation operation.
During training, given the ground truth emotional label $e^*$ of dialogue context, we employ negative log-likelihood as the loss function to conduct the parameter learning:
\begin{equation}
\setlength{\abovedisplayskip}{3pt}
    \mathcal{L}_{emo} = -\log(P_e(e^*|\mathcal{E})).
\label{eq:mle_loss}
\end{equation}
In addition, the obtained intermediate emotional representation $\mathbf{e}_p$ will be fed into the decoder as a crucial emotional feature to guild the empathetic response generation.

The final dialogue context representations $\mathbf{C}$ is the concatenation of the semantic context vectors $\mathbf{C_u}$ and emotional context vectors $\mathbf{C_e}$, i.e., $\mathbf{C} = [\mathbf{C_u};\mathbf{C_e}]$.

\paragraph{Empathetic Response Generation.}
\label{ssec:generator}
The predicted emotion signal $\mathbf{e}_p \in \mathbb{R}^{1 \times q}$ is firstly be transformed by a linear transformation into $\mathbf{e}'_p \in \mathbb{R}^{1 \times d}$. 
Then we concatenate $\mathbf{e}'_p$ with the embeddings of the decoder input tokens $[y_1,\ldots,y_{j-1}]$ into representations $\mathbf{E}^Y =[\mathbf{y}_{0}\ldots,\mathbf{y}_{j-1}]$ where $\mathbf{y}_0=\mathbf{e}'_p$.
We feed $\mathbf{E}^Y$ into the response generator.

The generator is built based on the Transformer layers as well.
For each Transformer decoder layer, the decoder inputs $\mathbf{E}^Y$ are first updated into new vector representations $\mathbf{Y}$. 
Then a multi-head cross-attention mechanism $\textrm{MH-CAtt}$~\cite{vaswani2017attention} derives a context vector from dialogue context:
\begin{eqnarray}
    \mathbf{D} &=& \mathbf{Y} + \mathbf{W}_m\textrm{MH-CAtt}(\mathbf{Y}, \mathbf{C}),\\
     \mathbf{\hat{D}} &=& \textrm{LayerNorm}(\mathbf{D}),\\
    \mathbf{\hat{Y}} &=& \textrm{LayerNorm}(\mathbf{\hat{D}} + \textrm{FFN}(\mathbf{\hat{D}})),
\end{eqnarray} 
where $\mathbf{\hat{Y}}=[\mathbf{\hat{y}}_1,\ldots,\mathbf{\hat{y}}_j]$.
The response generator yields the distribution over the vocabulary for the next $j$-th token:
\begin{equation} 
\setlength{\abovedisplayskip}{3pt}
\setlength{\belowdisplayskip}{3pt}
    p(y_j|y_{<j},\mathbf{C}) = \textrm{\text{softmax}}(\mathbf{W}_{o}\mathbf{\hat{y}}_j).
\end{equation}
As most dialogue generation tasks, we use standard maximum likelihood estimator (MLE) as the optimization objective:
\begin{equation}
\setlength{\abovedisplayskip}{3pt}
\setlength{\belowdisplayskip}{3pt}
     \mathcal{L}_{gen} = -\log p(y_j|y_{<j},\mathbf{C}).
\end{equation}
Finally, considering all the aforementioned components, we define a joint loss function as follows:
\begin{equation} 
\setlength{\abovedisplayskip}{3pt}
\setlength{\belowdisplayskip}{3pt}
     \mathcal{L}_g = \gamma_1\mathcal{L}_{emo} + \gamma_1\mathcal{L}_{gen},
\label{loss_g}
\end{equation}
where $\gamma_1,\gamma_2$ are hyper-parameters that control the weights of the two losses. We set $\gamma_1=\gamma_2=1$.
All the parameters are jointly trained in an end-to-end paradigm.

\subsection{Interactive Discriminators}
To evaluate whether the response is generated in an empathetic and context-consistent way, we introduce two discriminators to provide additional training signals for the empathetic generator.
A semantic discriminator measures the semantic distance from the generated response to the gold response. An emotional discriminator specifies whether the generated responses are empathetic enough.

Specially, the next utterance of response could serve as the user's implicit feedback and provide semantic and emotional guidance for target response~\cite{zhang2018exploring}.
Therefore, we regard the next utterance of response as user semantic feedback and its contained emotional words as user emotional feedback.
In the training procedure, we utilize user semantic feedback and emotional feedback to optimize the content and empathy ability of the empathetic generator, respectively.
Both the semantic discriminator and the emotional one are built on the convolutional neural network (CNN) based classifier, 
so we detail the semantic discriminator first for convenience.

\paragraph{Semantic Discriminator.}
First, we apply an LSTM encoder~\cite{hochreiter1997long} to respectively encode the generated response and gold response into hidden representations, i.e., $\textbf{d}_t^N$ and $\textbf{d}_t^P$. 
We regard $\textbf{d}_t^N$ as negative vectors and $\textbf{d}_t^P$ as positive vectors.

Thereafter, a two-dimensional convolutional layer~\cite{kalchbrenner2014convolutional} convolves the hidden vecor $\textbf{d}^*_t$ with multiple convolutional kernels of different widths, where $* \in \{N, P\}$.
Each kernel corresponds to a linguistic feature detector which extracts a specific pattern of multi-grained n-grams~\cite{kalchbrenner2014convolutional}.
A convolutional filter $\mathbf{W_s}$ maps hidden states in the receptive field to a single feature.
As we slide the filter across the negative or positive sequence, a sequence of new features $\mathbf{F}^*=[\mathbf{f}_1^*,\ldots, \mathbf{f}_k^*]$ is obtained:
\begin{equation}
\setlength{\abovedisplayskip}{3pt}
\setlength{\belowdisplayskip}{3pt}
    \mathbf{f}^{*}_t = \text{ReLU}(\mathbf{d}_t^{*} \otimes  \mathbf{W}_s + \mathbf{b}_s),
\end{equation}
where ReLU is activation function, $\otimes$ denotes the convolution operation, $\mathbf{W}_s\in \mathbb{R}^{d \times k}$ and $\mathbf{b}_s \in \mathbb{R}^k$ are learnable parameters in the convolutional filter. 
For each convolutional filter, the max-pooling layer takes the maximal value among the convolutional features $\mathbf{F}^*$ and results in a fixed-size vector $\mathbf{f}^*$.
Then we obtain the semantic classification result $\texttt{D}_{sem}(\textbf{d}_t^{*}) \in \mathbb{R}$ through interaction among $\mathbf{f}^*$, semantic feedback representation $\mathbf{d}^F$, and dialogue context vector $\mathbf{\tilde{x}}_0$~(we use emotional context vector $\mathbf{\tilde{w}}_0$ in emotional discriminator): 
\begin{equation}
\setlength{\abovedisplayskip}{3pt}
\setlength{\belowdisplayskip}{3pt}
    \texttt{D}_{sem}(\mathbf{d}^{*}) = \mathbf{W}_d \text{ReLU}(\mathbf{f}^{*}+\mathbf{d}^F+\mathbf{\tilde{x}}_0)+b_d,
\end{equation}
where $\mathbf{W}_d\in \mathbb{R}^{1\times d}$ and $b_d \in \mathbb{R}$ are model parameters. $\mathbf{d}^F$ is the last hidden state derived by the above-mentioned LSTM encoder.

Inspired by previous work~\cite{arjovsky2017wasserstein,gao2019product}, we minimize the Wasserstein-1 distance between distributions of positive samples and negative samples. Hence, the discriminator $\texttt{D}\in \mathcal{D}$ becomes a \textit{1-Lipschitz} function, where $\mathcal{D}$ is the set of \textit{1-Lipschitz} functions~\cite{gulrajani2017improved}.
In order to meet the \textit{1-Lipschitz} constraints of discriminator \texttt{D}, we incorporate a gradient penalty of the output of \texttt{D} with respect to its input into the discriminator objective function. 
The gradient penalty is sampled uniformly along a straight line between points sampled from the negative representations $\textbf{d}_t^N$ and the positive representations $\textbf{d}_t^P$.
Then the loss function of semantic discriminator is computed as follows:
\begin{eqnarray}
\setlength{\abovedisplayskip}{3pt}
\setlength{\belowdisplayskip}{3pt}
    &\mathbf{d}_t^{'} = \alpha \mathbf{d}_t^N + (1-\alpha) \mathbf{d}_t^P,  \\
    &\mathcal{L}^{sem}_d = \frac{1}{n}\sum_{t}^{n}\texttt{D}_{sem}(\mathbf{d}_t^N) -\texttt{D}_{sem}(\mathbf{d}_t^P) 
         + \beta(\| \nabla_{\mathbf{d}_t^{'}}\texttt{D}_{sem}(\mathbf{d}_t^{'}) \|_2 - 1)^2,
\label{loss_d}
\end{eqnarray}
where $\alpha \sim \texttt{U}[0,1]$ is a random value and $\beta$ denotes a coefficient of gradient penalty term.

\paragraph{Emotional Discriminator.}
The architecture of the emotional discriminator is the same as that of the semantic discriminator.
The main difference is that emotional discriminator conducts on the emotional words of the generated response, gold response, and user feedback~(i.e., user emotional feedback). 
We use $\mathcal{L}^{emo}_d$ to denote the loss function of emotional discriminator.

The total loss $\mathcal{L}_d$ of interactive discriminators is the loss summation of semantic discriminator $\mathcal{L}^{sem}_d$ and emotional discriminator $\mathcal{L}^{emo}_d$.
Meanwhile, we add the summation of $-\texttt{D}_{sem}(\mathbf{d}_t^N)$ and $-\texttt{D}_{emo}(\mathbf{d}_t^N)$ to $\mathcal{L}_g$ to facilitate empathetic generator with more empathy ability.

\subsection{Training}
At the beginning of the training, we use the maximum likelihood estimation (MLE) to pre-train empathetic generator (Eq.~\ref{loss_g}).
Since pre-trained discriminators is effective to help adjust the empathetic generator~\cite{yu2017seqgan}, we pre-train the interactive discriminators as well. 
After the pre-training, the empathetic generator and interactive discriminators are trained alternatively.

%% file: 4.experiment.tex
\section{Experiment}
\subsection{Dataset}
We evaluate EmpDG on the dataset \textsc{EmpatheticDialogues}~\cite{rashkin2019towards}, which is a large-scale multi-turn empathetic dialogue dataset collected on the Amazon Mechanical Turk platform, containing about 25k one-to-one open-domain conversations.
Specifically, Rashkin et al.~\shortcite{rashkin2019towards} pair two crowd-workers: a speaker and a listener. The speaker is asked to talk about the personal emotional situation. The listener infers the underlying emotion through what the speaker says and responds empathetically.
The dataset provides 32 evenly distributed emotion labels, which act as the coarse-grained dialogue-level emotions. 
We use the \textbf{NRC Emotion Lexicons}~(NRC)~\cite{mohammad2013crowdsourcing} to extract the emotional words in dialogue context to conduct fine-grained token-level emotions.
In order to supplement the language gap between the training data and NRC, all adjectives not included in NRC are extracted together with NRC emotional words.
We treat the dialogue context and fine-grained emotional context as system inputs. The target outputs are a coarse-grained emotion label and the listener's response.
For our model, we reserve the next utterance of the target response as user feedback in the training procedure. 
Finally, we obtain 20,724 dialogues in the training set, 2,972 in the validation set, and 2,713 in the testing set.

\subsection{Evaluation Methods}
\paragraph{Automatic Evaluation.}
Liu et al.~\shortcite{LiuLSNCP16} have verified BLEU might be improper to evaluate the conversation generation problem, as it correlates weakly with human judgements of the response quality; METEOR~\cite{BanerjeeL05} and ROUGE~\cite{lin2004rouge} have the same problem.
Therefore, following previous emotion-related studies~\cite{zhou2018emotional,rashkin2019towards,song2019generating,wei2019emotion}, we employ three evaluation metrics to automatically evaluate the performance of our EmpDG: Perplexity~\cite{SerbanSBCP15} measures the high-level general quality of the generation model;
Distinct-1 and Distinct-2~\cite{li2015diversity} measure the proportion of the distinct unigrams / bigrams in all the generated results to indicate diversity.
To evaluate the model at the emotional level, we adopt Emotion Accuracy as the agreement between the ground truth emotion labels and the predicted emotion labels by the empathetic generator.

\paragraph{Human Evaluation.}
To qualitatively examine model performance from both the content and the empathy perspectives, we also conduct widely-adopted human evaluations.
We randomly sample 100 dialogues and their corresponding generations from our model as well as the baselines. We recruit three professional annotators from a third-party company to evaluate the responses generated by different models.
All models are evaluated in terms of following 3 metrics: Empathy, Relevance and Fluency \cite{rashkin2019towards,LinMSXF19}.
Empathy measures whether the listener's responses show the understanding of the speaker's feelings;
Relevance evaluates whether the generated responses are on-topic with the dialogue context;
Fluency measures the grammatical correctness and readability of the generated responses.
Each metric is rated on five-scale, where 1, 3, and 5 indicate unacceptable, moderate, and excellent performance, respectively.

\subsection{Baselines}
We conducted extensive experiments to compare EmpDG against the following representative baselines: 
\begin{itemize}
    \item \textbf{Transformer}~\cite{vaswani2017attention}: A Transformer-based sequence to sequence model that is trained based on MLE loss.
    \item \textbf{EmoPrepend-1}~\cite{rashkin2019towards}: An extension of the Transformer model which incorporates an additional supervised emotion classifier. The whole model is jointly trained by optimizing both the classification and generation loss.
    \item \textbf{MoEL}~\cite{LinMSXF19}: Another extension of Transformer model which softly combines the response representations from different transformer decoders. Each decoder is optimized to focus on one type of emotion accordingly.
\end{itemize}

Additionally, to better analyze the influence of different components in our model, we also conducted ablation tests as follows:
\begin{itemize}
    \item \textbf{w/o G}: The EmpDG model without the multi-resolution emotion factors, where we only consider the coarse-grained dialogue-level emotions and semantic discriminator.
    \item \textbf{w/o D}: The EmpDG model without the interactive discriminators.
\end{itemize}

\subsection{Implementation Details}
We implement all models using Pytorch~\cite{paszke2017automatic}\footnote{Our code is available at \url{https://github.com/qtli/EmpDG}.} and optimize the models using Adam~\cite{kingma2014adam} with a mini-batch size of 16. 
We use pre-trained Glove vectors~\cite{pennington2014glove} to initialize the word embedding.
During the training of empathetic generator, the learning rate is initialled as 0.0001 and we vary the learning rate following Vaswani et al.~\shortcite{vaswani2017attention}.
Early stopping is applied when training. When inference, we set the maximum decoding step as 30.
All common hyperparameters are the same as the work in~\cite{LinMSXF19}.
During the interactive adversarial training, D-steps~(for two interactive discriminators) is set to 1 and G-steps~(for empathetic generator) is set to 5.
Hyper-parameter $\beta$ in interactive discriminators is set to 0.1.
Meanwhile, we employ the teacher-forcing technique from Li et al. \shortcite{li2017adversarial} to increase adversarial training efficiency.

\begin{table*}[!t]
\centering
\small
\resizebox{\columnwidth}{!}{
\begin{tabular}{l |c c c c | c  c  c}
\toprule
\multirow{1}{*}{\textbf{Models}} &
\multirow{1}{*}{\textbf{Accuracy}} &
\multirow{1}{*}{\textbf{Perplexity} $\downarrow$} &
\multirow{1}{*}{\textbf{Distinct-1}} &
\multirow{1}{*}{\textbf{Distinct-2}} &
\multirow{1}{*}{\textbf{Empathy}} &
\multirow{1}{*}{\textbf{Relevance}} &
\multirow{1}{*}{\textbf{Fluency}} 
\\
\midrule
Transformer & - & 33.91 & 1.17 & 4.70 & 3.15 & 3.42  & 3.64 \\
EmoPrepend-1 & 0.3328 & \textbf{33.35} & 1.06 & 4.29 &  3.29 & 3.59 & \textbf{3.71} \\
MoEL & 0.3200 & 33.58 & 1.38 & 4.66 & 3.47 & 3.88  & 3.68 \\
\midrule
EmpDG &  \textbf{0.3431} & 34.18 & \textbf{1.81} & \textbf{6.94}  & \textbf{3.58} & \textbf{3.91} & 3.67 \\
\bottomrule
\end{tabular}}
\caption{Evaluation results between baselines and our models. The first four metrics are automatic metrics, while the next three metrics are human metrics. \textbf{Bold face} indicates leading results in terms of the corresponding metric.}
\label{tab:auto_result}
\end{table*}

\subsection{Performance Comparisons}
\paragraph{Automatic Evaluation Results.}
As shown in Table \ref{tab:auto_result}, EmpDG achieves the highest scores for most metrics compared with other baselines. 
The noticeable improvement indicates the effectiveness of our  multi-resolution adversarial empathetic dialogue generation model in empathetic expression and response diversity.
Although the perplexity score of EmpDG is slightly worse than the EmoPrepend-1 due to the introduction of interactive discriminators, the other scores for EmpDG are obviously better than EmoPrepend-1.  
EmpPrepend-1 and MoEL have similar performance, as both of them only learn the coarse-grained dialogue-level emotional label to infer emotional situation and generate responses. 
Without emotion modelling, Transformer only generates fluent responses based on semantic mapping, but fail to express diverse responses.

We also perform an ablation study for better understanding the contributions of the main parts of our model. 
As shown in Table~\ref{tab:abla_tests}, after we remove the multi-resolution empathetic generator~(i.e., w/o G), both the emotion accuracy and distinct metrics performance become obviously worse, demonstrating the effectiveness of multi-resolution emotion factors in emotional understanding and model generation quality.
We also investigate the effect of removing interactive discriminators~(i.e., w/o D). We notice that the scores of distinct metrics decrease dramatically.
This makes sense because user feedback interacts with generated responses to facilitate empathetic generator optimization.
Therefore, applying interactive discriminators can bring performance improvement on appropriate emotional expressions.

\begin{table}[!t]
\centering
\resizebox{0.7\columnwidth}{!}{
\begin{tabular}{l |c | c | c | c}
  \toprule
  \multirow{1}{*}{\textbf{Models}} &
   \multirow{1}{*}{\textbf{Accuracy}} &
  \multirow{1}{*}{\textbf{Perplexity} $\downarrow$} &
  \multirow{1}{*}{\textbf{Distinct-1}} &
  \multirow{1}{*}{\textbf{Distinct-2}} 
  \\
  \midrule
  EmpDG & \textbf{0.3431} & 34.18 & \textbf{1.81} & \textbf{6.94} \\
  w/o G & 0.3281 & 33.94 & 1.27 & 5.21 \\
  w/o D  & 0.3347 & \textbf{32.66} & 1.00 & 3.89\\
  \bottomrule
\end{tabular}
}
\caption{Ablation test of different components.}
\label{tab:abla_tests}
\end{table}

\begin{table}[!t]
\centering
\resizebox{0.6\columnwidth}{!}{
\begin{tabular}{l |c | c | c}
  \toprule
  \multirow{1}{*}{\textbf{Models}} &
   \multirow{1}{*}{\textbf{Win}} &
  \multirow{1}{*}{\textbf{Loss}} &
  \multirow{1}{*}{\textbf{Tie}} 
  \\
  \midrule
  EmpDG vs Transformer & 40.2\% & 27.5\% & 32.3\% \\
  EmpDG vs EmoP  & 38.7\% & 29.4\% & 31.9\% \\
  EmpDG vs MoEL & 35.7\% & 31.2\% & 33.1\% \\
  \bottomrule
\end{tabular}
}
\caption{Result of human A/B test. Tests are conducted pairwise between EmpDG and baseline models.}
\label{tab:ab_result}
\end{table}

\paragraph{Human Evaluation Results.}
Table \ref{tab:auto_result} illustrates that EmpDG obtains the best performance on both Empathy and Relevance scores.
This suggests that the multi-resolution adversarial model helps capture implicit emotions, improve the topic consistency, and evoke more emotion perceptivity.
As the generated responses by Transformer are already fluent and grammatical, we observe there is no obvious difference among models in terms of Fluency. 
Additionally, we carried out pairwise response comparison to directly compare the dialogue quality gains in Table~\ref{tab:ab_result}.
The results confirm that the responses from EmpDG are more preferred by human judges.

\begin{table}[!t]
  \small
  \centering
  \begin{tabularx}{0.9\textwidth}{l | l }
  \toprule
  Semantic Context&\colorbox[rgb]{0.9921,0.9608,0.9020}{I}\colorbox[rgb]{0.9921,0.9608,0.9020}{am}\colorbox[rgb]{0.9921,0.9608,0.9020}{so}\colorbox[rgb]{1.0,0.8941,0.7686}{grateful}\colorbox[rgb]{0.9921,0.9608,0.9020}{for}\colorbox[rgb]{0.9921,0.9608,0.9020}{my}\colorbox[rgb]{1.0,0.8941,0.7686}{family}\colorbox[rgb]{0.9921,0.9608,0.9020}{due}\colorbox[rgb]{0.9921,0.9608,0.9020}{to}\colorbox[rgb]{0.9921,0.9608,0.9020}{an}\colorbox[rgb]{0.9921,0.9608,0.9020}{incident}\colorbox[rgb]{0.9921,0.9608,0.9020}{with}\colorbox[rgb]{0.9921,0.9608,0.9020}{my}\colorbox[rgb]{1.0,0.8941,0.7686}{friend}\colorbox[rgb]{1.0,0.8941,0.7686}{.} \\
  {\bf MoEL}  &  Yeah, it is good about being a while. \\
  \dashrule \dashrule \dashrule \dashrule & \dashrule\dashrule\dashrule\\
  Semantic Context&\colorbox[rgb]{0.9921,0.9608,0.9020}{I}\colorbox[rgb]{0.9921,0.9608,0.9020}{am}\colorbox[rgb]{0.9921,0.9608,0.9020}{so}\colorbox[rgb]{1.0, 0.74, 0.53}{grateful}\colorbox[rgb]{0.9921,0.9608,0.9020}{for}\colorbox[rgb]{0.9921,0.9608,0.9020}{my}\colorbox[rgb]{0.9921,0.9608,0.9020}{family}\colorbox[rgb]{0.9921,0.9608,0.9020}{due}\colorbox[rgb]{0.9921,0.9608,0.9020}{to}\colorbox[rgb]{0.9921,0.9608,0.9020}{an}\colorbox[rgb]{1.0,0.8941,0.7686}{incident}\colorbox[rgb]{0.9921,0.9608,0.9020}{with}\colorbox[rgb]{0.9921,0.9608,0.9020}{my}\colorbox[rgb]{0.9921,0.9608,0.9020}{friend}\colorbox[rgb]{0.9921,0.9608,0.9020}{.} \\
  Emotional Context& \colorbox[rgb]{1.0,0.8941,0.7686}{grateful}, \colorbox[rgb]{0.9921,0.9608,0.9020}{due},\colorbox[rgb]{1.0,0.8941,0.7686}{incident},\colorbox[rgb]{0.9921,0.9608,0.9020}{friend}\\
  {\bf EmpDG} &  That is \underline{good}. What kind of \underline{incident}?  \\
  User Feedback&  My friend lost her parents and siblings in a fire and their family was very close to \\ 
  & mine. So seeing her go through the grief made me realize how thankful I should be. \\
  \midrule
  Semantic Context&\colorbox[rgb]{0.9921,0.9608,0.9020}{I}\colorbox[rgb]{0.9921,0.9608,0.9020}{recently}\colorbox[rgb]{1.0,0.8941,0.7686}{had}\colorbox[rgb]{0.9921,0.9608,0.9020}{a}\colorbox[rgb]{1.0, 0.74, 0.53}{scary}\colorbox[rgb]{0.9921,0.9608,0.9020}{experience}\colorbox[rgb]{1.0,0.8941,0.7686}{.}\colorbox[rgb]{0.9921,0.9608,0.9020}{I}\colorbox[rgb]{0.9921,0.9608,0.9020}{heard}\colorbox[rgb]{0.9921,0.9608,0.9020}{gunshots} \colorbox[rgb]{0.9921,0.9608,0.9020}{on}\colorbox[rgb]{1.0,0.8941,0.7686}{my}\colorbox[rgb]{0.9921,0.9608,0.9020}{street}\colorbox[rgb]{1.0,0.8078,0.6235}{.} \\
  {\bf MoEL}  & Oh my god, I am not sure what I would be scary. \\
  \dashrule \dashrule  & \dashrule \\ 
  Semantic Context&\colorbox[rgb]{1.0,0.8941,0.7686}{I}\colorbox[rgb]{0.9921,0.9608,0.9020}{recently}\colorbox[rgb]{0.9921,0.9608,0.9020}{had}\colorbox[rgb]{0.9921,0.9608,0.9020}{a}\colorbox[rgb]{1.0,0.8941,0.7686}{scary}\colorbox[rgb]{1.0,0.8941,0.7686}{experience}\colorbox[rgb]{1.0,0.8941,0.7686}{.}\colorbox[rgb]{1.0,0.8941,0.7686}{I}\colorbox[rgb]{0.9921,0.9608,0.9020}{heard}\colorbox[rgb]{0.9921,0.9608,0.9020}{gunshots} \colorbox[rgb]{0.9921,0.9608,0.9020}{on}\colorbox[rgb]{0.9921,0.9608,0.9020}{my}\colorbox[rgb]{0.9921,0.9608,0.9020}{street}\colorbox[rgb]{0.9921,0.9608,0.9020}{.} \\
  Emotional Context& 
  \colorbox[rgb]{1.0,0.8549,0.72549}{scary},\colorbox[rgb]{0.9921,0.9608,0.9020}{experience},\colorbox[rgb]{0.9921,0.9608,0.9020}{heard},\colorbox[rgb]{0.9921,0.9608,0.9020}{street}\\
  {\bf EmpDG} &  Oh no, I am \underline{sorry} to hear that. What \underline{happened}? \\
  User Feedback& Everything turned out fine. I think somebody in the nearby neighborhoods may have \\
  & been hurt, though. \\
  \bottomrule
  \end{tabularx}
  \caption{Two visualization cases of the cross-attention weights in MoEL and EmpDG. Words with the underline in the generated responses of EmpDG contain context-consistent emotions.} 
  \label{tab:attn_dist}
  \vspace{-5mm}
  \end{table}

\subsection{Analysis of Emotion Interactions}
To gain an insight into how well the emotion is expressed in the generated responses, we portrait two examples illustrating the cross-attention weights~(between the encoder and decoder of the empathetic generator) of the dialogue context in the Table~\ref{tab:attn_dist}.
For the first visualization case, by giving additional attention to the multi-resolution emotional information, the response generated by EmpDG shows much more informative and empathetic, such as words ``good'' and ``incident''.
For the second visualization case, although MoEL puts the major attention on the emotional words, i.e., ``scary'', the content of its generated response is inappropriate, which is not enough to evoke much empathy. 
EmpDG can generate the more contentful response with better emotional expressions, e.g., ``sorry'' and ``what happened'' which are also relevant to the user feedback.
We can conclude that considering user feedback in adversarial training can teach the model to generate more empathetic responses.
In total, the effectiveness of the multi-resolution mechanism and interactive discriminators are proved to some extent.

\begin{table*}[h]
\small
\centering
\begin{tabular}{l|l}
\toprule
{\bf Emo.}&{\bf Hopeful}\\
{\bf S\_Cont.}&{$X_1$: \ My husband lost a job but I am hoping he can find a full time job soon.}\\
{\bf E\_Cont.}&{lost, job, hoping, find, full, time.}\\
\midrule
\textbf{Gold} &{\textbf{He will, I have faith.}} \\
Transformer & That is great! I am sure he will be fine. \\
EmoPrepend-1  & I am sorry to hear that. I hope he gets it.  \\
MoEL & That is a good news. I hope you get it for him.  \\
EmpDG & I am \underline{sorry} to hear that. I \underline{hope} he gets \underline{better}.\\
\midrule
\midrule
{\bf Emo.}&{\bf Grateful}\\
{\bf S\_Cont.}&{$X_1$: I joined a firm 6 months ago and then I got a promotion for junior manager.}\\
{}&{$X_2$: Congratulations. That sounds like the fast track.}\\
{}&{$X_3$: Yes, they thought that I have some extraordinary talent and that is the reason I got that position.}\\
{\bf E\_Cont.}&{joined, firm, ago, promotion, junior, manager, fast, track, thought, extraordinary, talent, reason, position.}\\
\midrule
\textbf{Gold} & \textbf{Cool. Maybe next year you will be a partner!} \\
Transfomer & I am sorry to hear that. \\
EmoPrepend-1  & That is great! i am glad you got it!  \\
MoEL & That is great, I bet it is hard to do that!  \\
EmpDG & \underline{Wow}, that is \underline{amazing}. I am sure they do \underline{great}. \\
\bottomrule
\end{tabular}
\caption{Generated responses from Transformer, EmoPrepend-1, MoEL, and our EmpDG in two different speaker's emotional states. S\_Cont indicates the dialogue semantic context and E\_Cont indicates the emotional context. $X_i$ represents the $i$-th utterance in semantic context. Tokens in \underline{underline} represent emotion-related words.
}
\label{tab:cases}
\end{table*}

\subsection{Case Study}
Table~\ref{tab:cases} shows two examples generated by EmpDG and other baseline models.
In the first case, EmpDG generates a coherent and informative response with a proper ``hopeful'' emotion by replying with words ``sorry'', ``hope'', ``gets'', ``better'', whereas baselines fail to understand the negative emotions or express the appropriate contents.
In the second case, EmpDG generates the most context-consistent response, which contains context-related words~(``do'', ``great'') and emotion-rated word~(``amazing'').

%% file: 5.conclusion.tex
\section{Conclusion}
In this paper, we propose an multi-resolution interactive empathetic dialogue model (EmpDG) to evoke more emotion perceptivity in dialogue generation.
Two components are proposed to improve the performance of empathetic response generation.
A multi-resolution empathetic generator combines coarse-grained dialogue-level and fine-grained token-level emotions to capture the nuanced emotional expressions in dialogue and evoke more emotion perceptivity.
Two interactive discriminators utilize user feedback as additional context to interact with the generated response and dialogue context to optimize the long-term goal of empathetic conversation generation.
Automatic and manual evaluation have shown that EmpDG can generate responses appropriate not only in content but also in empathy. 

There are several future directions for this setting. First, one of the potential extension of EmpDG would be incorporating it with external knowledge~(e.g., user profile or commonsense knowledge) to help emotion perceptivity.
Second, in our setting, the emotional feedback and semantic feedback are separated. It is promising to model the interaction between semantic and emotional feedback.
We leave these directions as future work.

\section*{Acknowledgements}
We want to thank our anonymous reviewers for their feedback.

This work was supported by the National Key R\&D Program of China with grant No. 2020YFB1406700, the Natural Science Foundation of China (61972234, 61902219, 61672324, 61672322, 62072279), the Key Scientific and Technological Innovation Program of Shandong Province (2019JZZY010129), the Tencent AI Lab Rhino-Bird Focused Research Program (JR201932), the Fundamental Research Funds of Shandong University, the Foundation of State Key Laboratory of Cognitive Intelligence, iFLYTEK, P.R. China (COGOSC-20190003).